\documentclass[11pt,letterpaper]{article}
\usepackage{emnlp2017}
\usepackage{times}
\usepackage{latexsym}
\usepackage{comment}
\usepackage{amssymb}
\usepackage{amsfonts}
\usepackage{mathtools}
\usepackage[utf8]{inputenc}
\usepackage{algorithm,algorithmic}
\usepackage{booktabs}

\emnlpfinalcopy

\def\emnlppaperid{***}

\renewcommand{\vec}[1]{\boldsymbol{#1}}
\newcommand{\norm}[1]{\left\lVert#1\right\rVert}

\renewcommand{\equationautorefname}{Eq.~\kern-0.2em}
\renewcommand{\sectionautorefname}{\S\kern-0.2em}
\renewcommand{\subsectionautorefname}{\S\kern-0.2em}
\renewcommand{\subsubsectionautorefname}{\S\kern-0.2em}

\title{The UMD Neural Machine Translation Systems \\ at WMT17 Bandit Learning Task}

\author{Amr Sharaf \and Shi Feng \and Khanh Nguyen \and Kiant{\'e} Brantley \and Hal Daum{\'e} III \\
    Department of Computer Science \\
    University of Maryland, College Park \\
    \small \texttt{\{amr,shifeng,kxnguyen,kdbrant,hal\}@cs.umd.edu}}

\date{}

\begin{document}

\maketitle

\begin{abstract}
    
We describe the University of Maryland machine translation systems submitted to
the WMT17 German-English Bandit Learning Task.
The task is to adapt a translation system to a new domain, using only \emph{bandit feedback}:
the system receives a German sentence to translate, produces an English sentence, and only gets a scalar score as feedback.
Targeting these two challenges (adaptation and bandit learning), we built a standard neural machine translation system
and extended it in two ways: (1)
robust reinforcement learning techniques to learn effectively
from the bandit feedback, and (2) domain adaptation using data selection from a large corpus of
parallel data.

\end{abstract}

\section{Introduction}

We describe the University of Maryland systems for bandit machine translation.
For the shared translation task of the EMNLP 2017's second conference on machine translation
(WMT17), we focused on the task of bandit machine translation.
This shared task was set up, consistent with \cite{kreutzer17bandit}, simultaneously as a bandit learning problem
\emph{and} a domain adaptation problem.
This raises the natural question:
can we combine these potentially complementary information sources?

To investigate this question, we started from a standard neural machine translation (NMT) setup \autoref{sec:neural_mt}\footnote{Our implementation is based on OpenNMT \cite{klein2017opennmt}, an  open-source toolkit for 
neural MT.}
, and then we:
\begin{enumerate}
	\item applied domain adaptation techniques by data selection~\cite{moore2010intelligent} 
	to the out-of-domain data, with the goals of filtering out harmful data and fine-tuning
	the training process to focus only on relevant sentences (\autoref{sec:da}).
	\item trained robust reinforcement learning algorithms that can effectively
	learn from bandit feedback (\autoref{sec:formulation}); this allows our model to ``test'' proposed 
	generalizations and adapt from the provided feedback signals.
\end{enumerate}

Tackling the problem of learning with bandit feedback is important because
neural machine translation systems, like other natural language processing 
technology, currently learn almost exclusively from labeled data for a specific 
domain. While this approach is useful, it cannot scale to a broad variety of
language and domains, as linguistic systems often cannot generalize well beyond 
their training data. 
Machine translation systems need to be able to learn to improve their 
performance from naturalistic interaction with users in addition to labeled data. 

Bandit feedback~\cite{robbins1985some} offers systems the opportunity to ``test'' proposed 
generalizations and receive feedback on their performance; particularly 
interesting are \emph{contextual} bandit
systems, which make predictions based on a given input context 
\cite{auer2002nonstochastic,langford2008epoch,beygelzimer2010contextual,dudik2011efficient}.
For example, a neural 
translation system trained on parliament proceedings often performs quite poorly 
at translating anything else. However, a translation system that is deployed to 
facilitate conversations between users might receive either explicit feedback 
(e.g. thumbs up/down) on its translations, or even implicit feedback, for example, 
the conversation partner asking for clarifications.
There has recently been a flurry of work specifically addressing the bandit 
structured prediction problem 
\cite{chang2015learning,sokolov2016learning,sokolov2016stochastic}, of which 
machine translation is a special case.

Because this task is---at it's core---a domain adaptation problem (for which a 
bandit learning signal is available to ``help''),
we also explored the use of standard domain adaptation techniques.
We make a strong assumption that a sizable amount of \emph{monolingual, source 
language} data is available
\emph{before} bandit feedback begins.\footnote{This raises a natural question: 
in the cases where this assumption is unreasonable, could we do adaptation online?}
We believe that in many realistic settings, one can at least get some amount of 
unlabeled data to begin with (we consider $40k$ sentences).
Using this monolingual data, we use data selection on a large corpus of parallel 
out-of-domain data (Europarl, NewsCommentary, CommonCrawl, Rapid) to seed an 
initial translation model.

Overall, the results support the following conclusions (\autoref{sec:exp}), 
based on the limited setting of one new domain and one language pair:
\begin{enumerate}
\item data selection for domain adaptation alone improves translation quality by 
  about $1.5$ \textsc{Bleu} points.
\item on \emph{top} of the domain adaptation, reinforcement learning 
  (which requires exploration) leads to an \emph{initial} degradation of about 
  $3$ \textsc{Bleu} points, which is recovered (on development data) after 
  approximately $40k$ sentences of bandit feedback.\footnote{Unfortunately, due 
    to our implementation bug, our evaluation of the test server is incomplete 
    for the reinforcement learning setting; see \autoref{sec:exp:rl} for a discussion.}
\end{enumerate}

One limitation of our current setup is that we used bandit feedback on 
development data to train a ``critic'' function for
our reinforcement learning implementation, which, in the worst case, means that 
our results over-estimate performance on the first $120k$ examples 
(more details in \autoref{sec:exp:rl}).





\section{Neural MT architecture}
\label{sec:neural_mt}

We closely follow \newcite{luong2015effective} for the structure of our neural
machine translation (NMT) systems.
Our NMT model consists of an encoder and a decoder, 
each of which is a recurrent neural network (RNN).
We use a bi-directionaral RNN as the encoder and a uni-directional RNN as the decoder. 
The model directly estimates the posterior distribution 
$P_{\theta}(\vec y \mid \vec x)$ of translating a source sentence $\vec x = (x_1, \cdots, x_n)$ to
a target sentence $\vec y = (y_1, \cdots, y_m)$:
\begin{align}
  P_{\theta}(\vec y \mid \vec x) = \prod_{t = 1}^m P_{\theta}(y_t \mid \vec y_{<t}, \vec x)
\end{align}
where $\vec y_{<t}$ are all tokens in the target sentence prior to $y_t$. 

Each local distribution $P_{\theta}(y \mid \vec y_{<t}, \vec x)$ is modeled as a multinomial
distribution over the target language vocabulary. We represent this as a linear
transformation followed by a softmax function on the decoder's output vector 
$\tilde{\vec h}_t^{dec}$: 
\begin{align}
  P_{\theta}(y \mid \vec y_{<t}, \vec x) &= \mathrm{softmax}(\vec W_s \ \tilde{\vec h}_t^{dec} ; \tau) \\
  \tilde{\vec h}_{t}^{dec} &= \tanh (\vec W_o [\vec h_t^{dec}; \vec c_t]) \\
  \vec c_t &= \mathrm{attend}(\vec h_{1:n}^{enc}, \vec h_t^{dec})
\label{eqn:softmax}
\end{align} where 
$[.;.]$ is the concatenation of two vectors,
$\mathrm{attend}(.,.)$ is an attention mechanism,
\footnote{We use the ``concat'' mechanism in \cite{luong2015effective}.},
$\tau$ is the temperature hyperparameter of the softmax function, 
$\vec h^{enc}$ and $\vec h^{dec}$ are the hidden vectors generated by the encoder
and the decoder, respectively.

During training, the encoder first encodes $\vec x$ to a continuous
vector $\vec \Phi(\vec x)$, which is used as the initial hidden vector for the decoder. 
The decoder performs RNN updates to produce a sequence of hidden vectors:
\begin{equation}
\begin{split}
  \vec h_0^{dec} &= \vec \Phi(\vec x) \\
  \vec h_t^{dec} &= f_{\theta} 
  \left( \vec h_{t - 1}^{dec}, \left[ \tilde{\vec h}_{t-1}^{dec}; \vec e(y_t) \right] \right)
\end{split}
\label{eqn:decoder}
\end{equation} where $\vec e(.)$ is a word embedding lookup operation, 
$f_{\theta}$ is an LSTM cell.
\footnote{Feeding $\tilde{\vec h}_t^{dec}$ to the next step is ``input feeding.''}

At prediction time, the ground-truth token $y_t$ in \autoref{eqn:decoder} 
is replaced by the model's own prediction $\hat{y}_t$:
\begin{equation}
  \hat{y}_t = \arg \max_y P_{\theta}(y \mid \hat{\vec y}_{<t}, \vec x)
\end{equation}

In a supervised learning framework, an NMT model is typically trained under the 
maximum log-likelihood objective:
\begin{equation}
\begin{split}
  \mathcal{L}_{sup}(\theta) &= 
  \mathbb{E}_{(\vec x, \vec y) \sim D_{\textrm{tr}}} 
  \left[ \log P_{\theta} \left( y \mid \vec x \right) \right]
\end{split}
\end{equation} where $D_{\textrm{tr}}$ is the training set.  


However, this learning framework is not applicable to our problem since 
reference translations are not available.

\section{Reinforcement Learning} \label{sec:formulation}

The translation process of an NMT model can be viewed as a Markov decision process
operating on a continuous state space. 
The states are the hidden vectors $\vec h_t^{dec}$ generated by the
decoder. 
The action space is the target language's vocabulary.

\subsection{Markov decision process formulation}

To generate a translation from a source sentence $\vec x$, an NMT model commences at an 
initial state 
$\vec h_0^{dec}$, which is a representation of $\vec x$ computed by the encoder.
At time step $t > 0$, the model decides the next action to take 
by defining a stochastic policy $P_{\theta}(y_t \mid \vec y_{<t}, \vec x)$,
which is directly parametrized by the parameters $\theta$ of the model. 
This policy takes the previous state $\vec h_{t - 1}^{dec}$ as input and produces a
probability distribution over all actions (words in the target vocabulary).
The next action $\hat{y}_t$ is chosen either by taking $\arg\max$ or sampling 
from this policy.
The encoder computes the current state $\vec h_t^{dec}$ by applying an RNN update on 
the previous state $\vec h_{t - 1}^{dec}$ and the next action taken $\hat{y}_t$ 
(\autoref{eqn:decoder}).

The objective of bandit NMT is to find a policy that maximizes the expected
quality of translations sampled from the model's policy:
\begin{equation}
  \mathcal{L}_{pg}(\theta) = 
  \mathbb{E}_{\substack{\vec x \sim D_{\textrm{tr}}\\\hat{\vec y} \sim P_{\theta}(\vec y \mid \vec x)}}
  \Big[ R(\hat{\vec y}, \vec x)  \Big]
  \label{eqn:reward_max}
\end{equation} where $R$ is a reward function that returns a score in $[0, 1]$
reflecting the quality of the input translation.

We optimize this objective function by policy gradient methods. 
The gradient of the objective in \autoref{eqn:reward_max} with respect to $\theta$ is:
\footnote{For notation brevity, we omit $\vec x$ from this equation. The expectations
are also taken over all given $\vec x$.}
\begin{align}
\label{eqn:pg_grad}
&  \nabla_{\theta} \mathcal{L}_{pg}(\theta) = 
  \mathbb{E}_{\hat{\vec y} \sim P(\cdot)}
  \left[ R(\hat{\vec y}) \nabla_{\theta} \log P_{\theta}(\hat{\vec y})  \right]  \\
  &= \sum_{t = 1}^m \mathbb{E}_{\substack{\hat{y}_t \sim\\P(\cdot \mid \hat{\vec y}_{<t})}}
  \Big[
    R(\hat{\vec y})
    \nabla_{\theta} \log P_{\theta}(\hat{y}_t \mid \hat{\vec y}_{<t})
   \Big] \nonumber
\end{align} 


\subsection{Advantage Actor-Critic} \label{sec:a2c}

\begin{algorithm}
  \caption{The A2C algorithm for NMT.}
  \label{alg:a2c}
\begin{algorithmic}[1]
  \small
  \FOR{$k = 0 \cdots K$}
  \STATE receive a source sentence $\vec x$
  \STATE sample a translation: $\hat{\vec y} \sim P_{\theta}(\vec y \mid \vec x)$
  \STATE receive reward $R(\hat{\vec y}, \vec x)$
  \STATE update the NMT model using the gradient in \autoref{eqn:pg_grad}
  \STATE update the critic model using the gradient in \autoref{eqn:critic_grad}
  \ENDFOR
\end{algorithmic}
\end{algorithm}

We follow the approach of the advantage actor-critic (A2C) algorithm 
\cite{mnih2016asynchronous}, 
which combines the REINFORCE algorithm \cite{williams1992simple} with actor-critic.
The algorithm approximates the gradient in \autoref{eqn:pg_grad} 
by a single-point sample and normalize the rewards by $V$ values to reduce
variance:
\begin{align}
  \nabla_{\theta} \mathcal{L}_{pg}(\theta) & \approx 
  \sum_{t = 1}^m \nabla_{\theta} 
  \log P_{\theta}(\hat{y}_t \mid \hat{\vec y}_{<t}, \vec x) \bar{R}_t(\hat{\vec y}_{<t}, \vec x) \nonumber\\
  \textrm{with } & \bar{R}_t(\hat{\vec y}_{<t}, \vec x) \equiv R(\hat{\vec y},
  \vec x) - V(\hat{\vec y}_{<t}, \vec x) 
\label{eqn:a2c_grad}
\end{align} where $\hat{y}_t \sim P(\cdot \mid \hat{\vec y}_{<t}, \vec x)$ and
$V(\hat{\vec y}_{<t}, \vec x) = \mathbb{E} \left[ R(\hat{\vec y}, \vec x) \mid \hat{\vec y}_{<t}, \vec x \right]$ 
is a baseline that estimates the expected future reward given $\vec x$ and $\hat{\vec y}_{<t}$. 

We train a critic model $V_{\omega}$ to estimate the $V$ values.
This model is an attention-based encoder-decoder model that encodes a source 
sentence $\vec x$ and decodes a predicted translation $\hat{\vec y}$.
At time step $t$, it computes
  $V_{\omega}(\hat{\vec y}_{<t}, \vec x) = \vec W_o \ \tilde{\vec h}_t^{dec}  $
  where $\tilde{\vec h}_t^{dec}$ is the hidden state of the RNN decoder,
and $\vec W_o$ is a matrix that transforms a vector into a scalar.
\footnote{We abuse the notation $\tilde{\vec h}^{dec}$ to denote the decoder
output. But since the translation model and the critic model do not share
parameters, their decoder outputs are distinct.}

The critic model is trained to minimize the MSE between its estimates and the true
values:
\begin{equation}
  \mathcal{L}_{crt}(\omega) = \mathbb{E}_{\vec x \sim D_{\textrm{tr}}} \left[
 \sum_{t = 1}^m \norm{ R(\hat{\vec y}, \vec x) - V_{\omega}(\hat{\vec y}_{<t}, \vec x) }^2 \right]
 \label{eqn:vnet}
\end{equation} 

Given a fixed $\vec x$, the gradient with respect to $\omega$ of this objective is:
\begin{equation}
  \nabla_{\omega} \mathcal{L}_{crt}(\omega) = 
 \sum_{t = 1}^m \left[ R(\hat{\vec y}) - 
 V_{\omega}(\hat{\vec y}_{<t}) \right]
                      \nabla_{\omega} V_{\omega}(\hat{\vec y}_{<t})
 \label{eqn:critic_grad}
\end{equation}

Algorithm~\autoref{alg:a2c} describes our algorithm. 
For each $\vec x$, we draw a single sample $\hat{\vec y}$ from the NMT model, 
which is used for both estimating the gradient of the NMT model 
(\autoref{eqn:a2c_grad}) and the gradient of the critic model (\autoref{eqn:critic_grad}).
We update the NMT model and the critic model simultaneously.


\section{Domain Adaptation} \label{sec:da}

We performed domain adaptation by choosing the best out-of-domain parallel data for training using~\citet{moore2010intelligent} cross-entropy based data selection technique.

\subsection*{Cross-Entropy Difference}
The Moore and Lewis method uses the cross-entropy difference $H_{I}(s)$ -
$H_{O}(s)$ for scoring a given sentence $s$, based on an in-domain language model
$LM_{I}$ and an out-of-domain language model $LM_{O}$
\cite{moore2010intelligent}. We trained $LM_{O}$ using the German-English
Europarl, NewsCommentary, CommonCrawl and Rapid (i.e. out-of-domain) data sets
and $LM_{I}$ using the e-commerce domain data provided by Amazon. After training
both language models, we follow Moore and Lewis method by applying the
cross-entropy difference to score each sentence in the out-of-domain data. The
cross-entropy is mathematically defined as:

\begin{align*}
    H(W) = - \frac{1}{n} \sum_{i=1}^{n}{\log P_{LM} (w_i | w_1, \cdots, w_{i-1})} 
\end{align*}

where $P_{LM}$ is the probability of a LM for the word sequence $W$ and $w_1, \cdots, w_{i-1}$ represents the history of the word $w_i$.

Sentences with the lowest cross-entropy difference scores are the most relevant
because they are the more similar to the in-domain data and less similar to the
average of the out-of-domain data. Using this criteria, the top $n$
out-of-domain sentences are used to create the training set $D_{tr}$. In this
work we consider various $n$ sizes, selecting the $n$ that provides the best
performance on the validation set.

\section{Experiments} \label{sec:exp}
This section describes the experiments we conducted in attempt to assess the 
challenges posed by bandit machine translation and our exploration of efficient
algorithms to improve machine translation systems using bandit feedback.

As explained in previous sections, this task requires performing domain
adaptation for machine translation through bandit feedback. With this in mind,
we experimented with two types of models: simple domain adaptation without using
the feedbacks, and reinforcement learning models that leverage the feedbacks.
In the following sections, we explain how we train the regular NMT model, how we
select training data for domain adaptation, and how we use reinforcement
learning to leverage the bandit feedbacks.

We trained our systems using the out-of-domain parallel data restricted by the 
shared task. The entire out-of-domain dataset contains 4.5 millions parallel 
German-English sentences from Europarl, NewsCommentary,
CommonCrawl and Rapid data for the News Translation (constrained) task.  Our NMT
model is based on OpenNMT's \cite{klein2017opennmt} PyTorch implementation of
attention-based encoder-decoder model. 
We extended their implementation and added our implementation of the A2C algorithm. 
Details of the model configuration and training hyperparameters
are listed in Table~\ref{table:model_details}.

\begin{table}[t]
  \centering
    \begin{tabular}{ll}
        \toprule
        \textbf{Word embedding size} & 500 \\
        \textbf{Hidden vector size} & 500 \\
        \textbf{Number of LSTM layers} & 2 \\
        \textbf{Batch size} & 64 \\
        \textbf{Epochs} & 13 \\
        \textbf{Optimizer} & SGD \\
        \textbf{Initial learning rate} & 1 \\
        \textbf{Dropout} & 0.3 \\
        \textbf{BPE size} & 20000 \\
        \textbf{Vocab size} & $\sim$25k (*) \\\bottomrule
    \end{tabular}
    \caption{NMT model's training hyperparameters. (*) with BPE we no longer
        need to prune the vocabulary, and the exact size depends on the training
        data.}
    \label{table:model_details}
\end{table}


\subsection{Subword Unit for Neural Machine Translation}

Neural machine translation (NMT) relies on first mapping each word into the vector space,
and traditionally we have a word vector corresponding to each word in a fixed vocabulary.
Due to the data scarcity, it's hard for the system to learn high quality representations
for rare words. To address this problem, with the goal of open vocabulary NMT,
~\citet{SennrichHB15} proposed to learn subword units and perform translation on a subword level.
We incorporated this approach in our system as a preprocessing step.
We generate the so-called byte-pair encoding (BPE), which is a mapping from words to subword units,
on the whole training set (WMT15), for both the source and target languages.
The same mapping is used for all the training sets in our system. 
After the translation, we do an extra post-processing step to convert the 
target language subword units back to words.
With BPE, the vocabulary size is reduced dramatically and we no longer need to
prune the vocabularies.
We find this approach to be very helpful and use it for all our systems.




\subsection{Domain Adaptation}

As explained in Section~\ref{sec:da}, we use the data selection method of
\cite{moore2010intelligent} for domain adaptation.  We use the kenlm toolkit~\cite{heafield2011kenlm} to
build all the language models used for the data selection. We train 4-gram
language models. For computing the cross-entropy similarity scores, we use the
XenC~\cite{rousseau13} open source data selection tool. We use the mono-lingual
data selection mode of XenC on the in-domain and out-of-domain source sentences.

We have two parameters in this data selection process: the size of in-domain
dataset that is used for training the in-domain language model, and the size of
the out-of-domain training data that we select. We experimented with different
configurations and the results on the development server are listed in
Table~\ref{table:adapt_dev}. For
obtaining the in-domain data, we pre-fetch the source sentences from development
and training servers.  For the training server, we do not have enough keys to
test all combinations, so we picked several configurations and for each
sentence, we select randomly a system to translate it. In addition, we also compare with and without beam
search. The purpose for this is to provide another comparable baseline for the
later reinforcement learning model, for which beam search cannot be used. Thus, 
the domain adaptation system that we submit to the training server is the
uniformly random combination of 6 systems, and their individual average
\textsc{bleu} scores are listed in Table~\ref{table:adapt_train}.  

\begin{table}[t]
    \centering
    \begin{tabular}{rcccc}
      \toprule
      & \multicolumn{3}{c}{\textbf{in-domain size}}\\
\textbf{o.o.d.\%}             & \textbf{40k} & \textbf{200k} & \textbf{800k} \\\midrule
        \textbf{10\%} & 18.50 & 18.57 & 18.85 \\
        \textbf{20\%} & 19.56 & 19.41 & 19.23 \\
        \textbf{30\%} & 19.54 & \textbf{20.16} & 19.11 \\
        \textbf{40\%} & \textbf{19.58} & 19.37 & 19.36 \\
        \textbf{60\%} & 18.88 & 18.81 & \textbf{19.59} \\
        \textbf{85\%} & 19.12 & 18.69 & 18.26 \\\midrule
        \textbf{(*) 100\%} & 18.70 & 18.70 & 18.70 \\\bottomrule
    \end{tabular}
    \caption{average \textsc{bleu} scores of domain adaptation systems on the
        development server with different combinations of in-domain size
        (x-axis) and the percentage of out-of-domain data selected (y-axis). (*)
        we show the \textsc{bleu} score of using all the out-of-domain data, do
    data selection performed for this row.}
    \label{table:adapt_dev}
\end{table}

\begin{table}[t]
    \centering
    \begin{tabular}{rrcc}
        \toprule
        \textbf{i.d. size}  & \textbf{o.o.d. \%} & \textbf{beam=1} & \textbf{beam=5} \\\midrule
        \textbf{0} & 100\%   & 18.07 & 18.65 (+0.58) \\
        \textbf{40k}  & 40\% & 18.77 & 19.51 (+0.74) \\
        \textbf{200k} & 30\% & \textbf{19.67} & \textbf{20.13  (+0.46)} \\ \bottomrule
    \end{tabular}
    \caption{Average \textsc{Bleu} scores of domain adaptation systems on the training
    server with different combinations of in-domain size, 
    out-of-domain percentage, beam size, and the corresponding \textsc{Bleu} scores.}
    \label{table:adapt_train}
\end{table}

It can be seen from these results that most configurations of data selection
improve the overall \textsc{Bleu} score. The model without data selection
achieves $18.70$ \textsc{Bleu} on the development server, while the best data
selection configurations achieves $20.16$, while on the training server the
scores are $18.65$ without data selection and $20.13$ with. It can also be seen
from Table~\ref{table:adapt_train} that beam search does help with improving the
\textsc{Bleu} score.

\subsection{Reinforcement Learning Results}
\label{sec:exp:rl}


While translating with the domain adaptation models to the development server,
we collect 320,000 triples of (source sentence, translation, feedback) from 8
submitted systems. We use these triples to pre-train the critic in the A2C algorithm. 
We use
the same pre-trained critic for all A2C-trained systems. The critic for each
model is then updated jointly with the actor respectively.  We use Adam
\cite{kingma14adam}
with learning rate of $10^{-4}$ to update the both the translation model and
the critic model. We do not use dropout
\cite{srivastava2014dropout} during training with A2C as it makes learning 
less stable.

We note that there are some drawbacks when using the A2C algorithm when it comes to
generating translations.  Normally we generate translations by greedy decoding,
which means at each time step we pick the word with the highest probability from
the distribution produced by the model. But with A2C, we need to sample
from the distribution of words to ensure exploration. As a direct consequence,
it is not clear how to apply beam search for A2C (and for policy gradient methods in 
general). To control the
trade-off between exploration and exploitation, we use the temperature
hyperparameter $\tau$ in the softmax function. In our experiments $\tau$ is
set to $\frac{2}{3}$, which produces a more peaky distribution and makes the 
model explore less.

It is best to have batching during bandit training for stability. Due to the limitation
of the submission servers, that is, we only get the single reward feedback each
time, we had to devise a method for batching for the feedback from the server.
We cache the rewards until we reach the batch size, then do a batch update.
However, due to some bugs in the implementation of this method, some sentences
are not submitted in the correct order. And at some test points on the training
server the scores are near or equal to zero.

\begin{figure}[t]
    \includegraphics[width=0.5\textwidth]{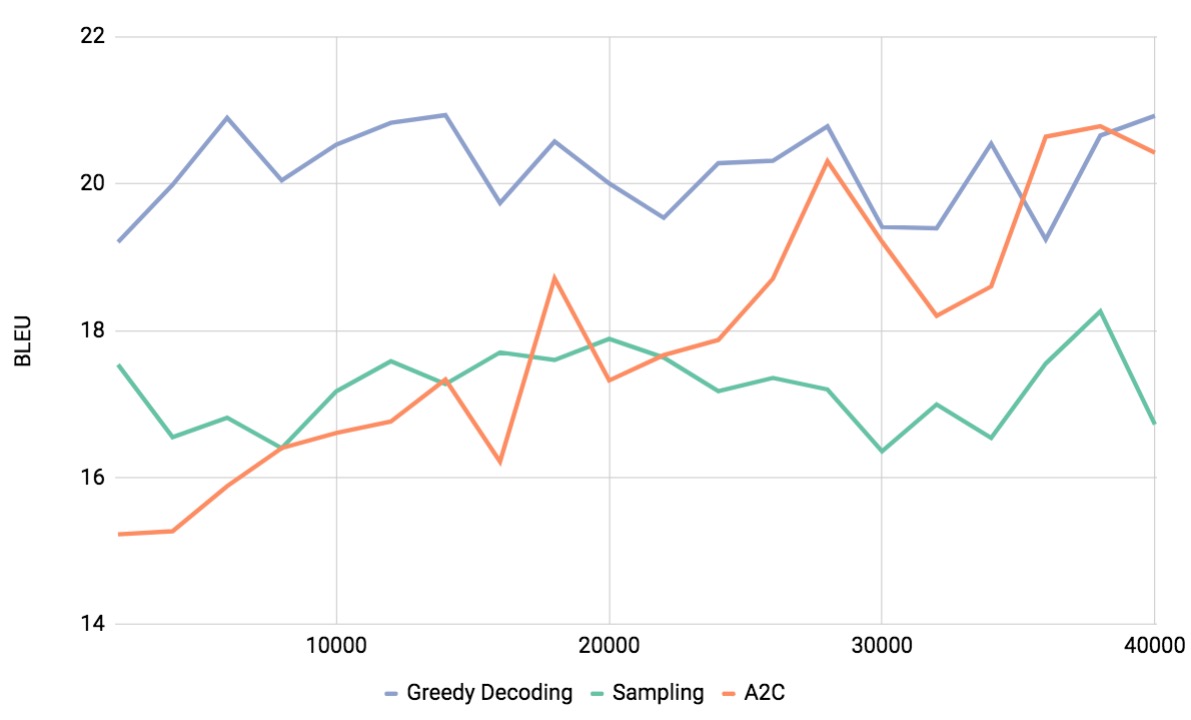}
    \caption{Comparing sampling, greedy decoding, and the A2C algorithm on the
      development data. Lines show average \textsc{BLEU} scores of every 2000
      consecutive sentences.}
    \label{figure:reinforce_dev}
\end{figure}

In Figure~\ref{figure:reinforce_dev} we present some results from the
development server.
We use a data selection model (200k in-domain data, 30\%
out-of-domain training data) as the baseline translation model, upon which we
use the A2C algorithm to improve further.
From this model, we generate translations with
both sampling and greedy decoding to see how much the exploration required by
the A2C algorithm hurts the performance. Figure~\ref{figure:reinforce_dev} shows the
average \textsc{BLEU} score of every 2000 sentences from the development server.
A2C loses at the beginning because of exploration, and catches up as it
sees more examples. Using sampling instead of greedy decoding, but exploration
eventually improves the model.

\section{Conclusion}
We present the University of Maryland neural machine translation systems 
for the WMT17 bandit MT shared task. 
We employ two approaches: 
out-of-domain data selection and reinforcement learning.
Experiments show that the best performance is achieved with a model pre-trained
with only one-third of the available out-of-domain data.
When applying reinforcement learning to further improve this model with bandit 
feedback, the model performance degrades initially due to exploration but gradually
improves over time.
Future work is to determine if reinforcement learning is more effective on a
 larger bandit learning dataset.

\section*{Acknowledgements}
The authors thank the anonymous reviewers for many helpful comments. We would
like to thank the task organizers: Pavel Danchenko, Hagen Fuerstenau, Julia 
Kreutzer,  Stefan Riezler,  Artem Sokolov, Kellen Sunderland,  and  Witold 
Szymaniak for organizing the task and for their help throughout the process. 

This work was supported by NSF grants IIS-1320538 and IIS-1618193, as well as 
an Amazon Research Award and LTS grant DO-0032. Any opinions, findings, 
conclusions, or recommendations expressed here are those of the authors and 
do not necessarily reflect the view of the sponsor(s).

\bibliography{emnlp2017}
\bibliographystyle{emnlp_natbib}

\end{document}